\documentclass[10pt,twocolumn,letterpaper]{article}

\usepackage{iccv}
\usepackage{times}
\usepackage{epsfig}
\usepackage{graphicx}
\usepackage{amsmath}
\usepackage{amssymb}
\usepackage{times}
\usepackage{epsfig}
\usepackage{graphicx}
\usepackage{amsmath}
\usepackage{amssymb}
\usepackage{subcaption}
\usepackage{epsfig}
\usepackage{booktabs}
\usepackage{multirow}

\usepackage[breaklinks=true,bookmarks=false]{hyperref}

\iccvfinalcopy 

\def\iccvPaperID{****} 
\def\httilde{\mbox{\tt\raisebox{-.5ex}{\symbol{126}}}}

\ificcvfinal\fi

\begin{document}

\def\httilde{\mbox{\tt\raisebox{-.5ex}{\symbol{126}}}}


\title{DeblurGAN-v2: Deblurring (Orders-of-Magnitude) Faster and Better}
\author{Orest Kupyn\textsuperscript{1, 3}, Tetiana Martyniuk\textsuperscript{1}, Junru Wu\textsuperscript{2}, Zhangyang Wang\textsuperscript{2} \vspace{2mm} \\
  \textsuperscript{1} Ukrainian Catholic University, Lviv, Ukraine; \qquad \textsuperscript{3} SoftServe, Lviv, Ukraine \\
  \texttt{\{kupyn, t.martynyuk\}@ucu.edu.ua} \vspace{2mm} \\
  \textsuperscript{2} Department of Computer Science and Engineering, Texas A{\&}M University \\
  \texttt{\{sandboxmaster,  atlaswang\}@tamu.edu}
  }

\maketitle
\begin{abstract}
\vspace{-0.5em}
We present a new end-to-end generative adversarial network (GAN) for single image motion deblurring, named DeblurGAN-v2, which considerably boosts state-of-the-art deblurring efficiency, quality, and flexibility. DeblurGAN-v2 is based on a relativistic conditional GAN with a double-scale discriminator. For the first time, we introduce the Feature Pyramid Network into deblurring, as a core building block in the generator of DeblurGAN-v2. It can flexibly work with a wide range of backbones, to navigate the balance between performance and efficiency. The plug-in of sophisticated backbones (e.g., Inception-ResNet-v2) can lead to solid state-of-the-art deblurring. Meanwhile, with light-weight backbones (e.g., MobileNet and its variants), DeblurGAN-v2 reaches 10-100 times faster than the nearest competitors, while maintaining close to state-of-the-art results, implying the option of real-time video deblurring. We demonstrate that DeblurGAN-v2 obtains very competitive performance on several popular benchmarks, in terms of deblurring quality (both objective and subjective), as well as efficiency. Besides, we show the architecture to be effective for general image restoration tasks too. Our codes, models and data are available at: \url{https://github.com/KupynOrest/DeblurGANv2}.
\end{abstract}
\vspace{-1em}
\section{Introduction}
\label{s:intro}

\newcommand{\ra}[1]{\renewcommand{\arraystretch}{#1}}
\begin{figure}[htb]
   \vspace{-2.2em}
\centering
  \includegraphics[width=\linewidth]{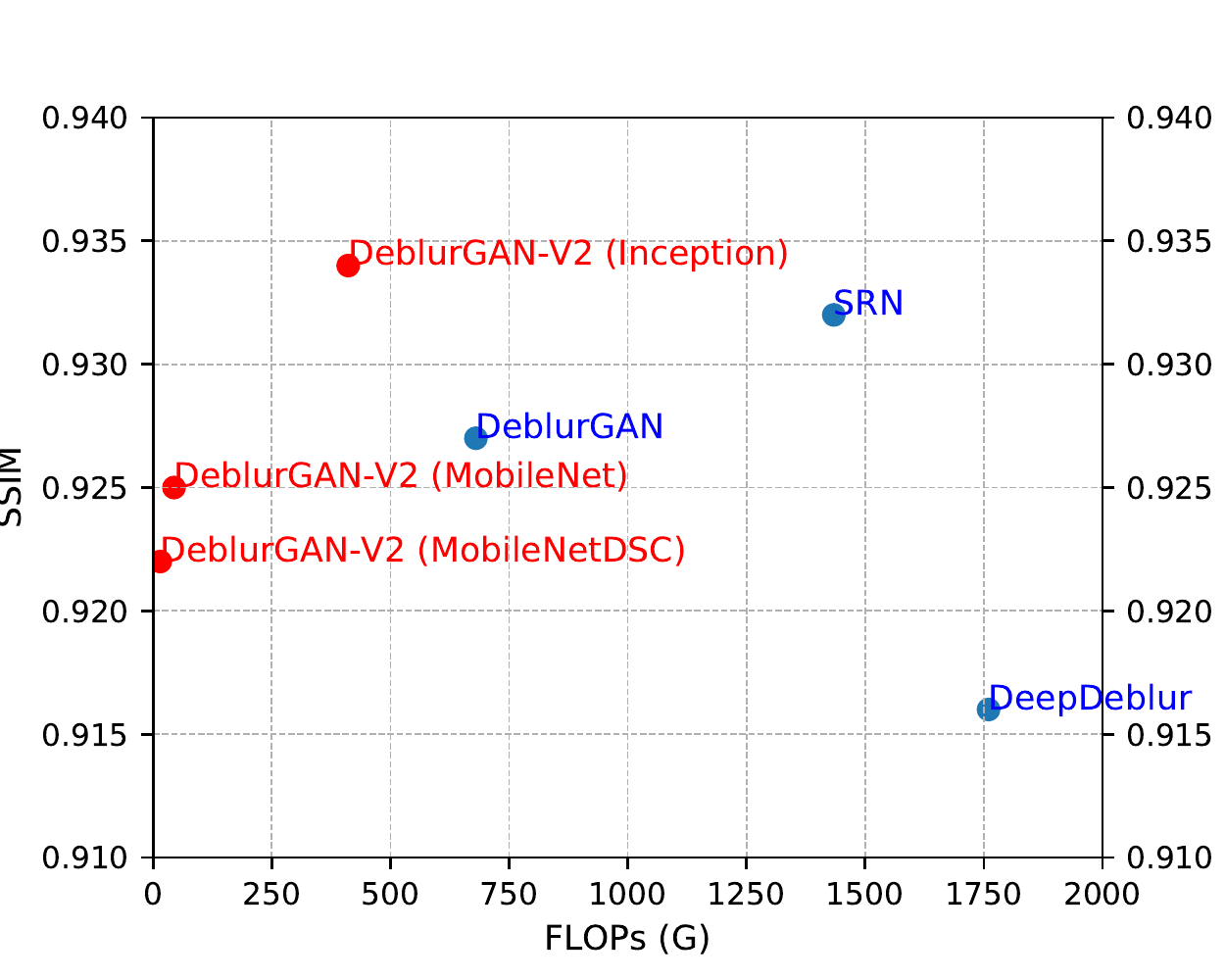} \\
     \vspace{-1em}
   \caption{The SSIM-FLOPs trade-off plot on the GoPRO dataset. Compared to three state-of-the-art competitors (in blue): DeblurGAN \cite{kupyn2018deblurgan}, DeepDeblur \cite{Nah2016DeepDeblurring} and Scale-Recurrent Network (SRN) \cite{tao2018scale}, DeblurGAN-v2 models (with different backbones, in red) are shown to achieve superior or comparable quality, and are much more efficient.}
   \vspace{-1em}
 \label{fig:cover}
\end{figure}

This paper focuses on the challenging setting of single-image blind motion deblurring. Motions blurs are commonly found from photos taken by hand-held cameras, or low-frame-rate videos containing moving objects. Blurs degrade the human perceptual quality, and challenge subsequent computer vision analytics. The real-world blurs typically have unknown and spatially varying blur kernels, 
and are further complicated by noise and other artifacts. 




The recent prosperity of deep learning has led to significant progress in the image restoration field \cite{wang2016d3,liu2017robust}. Specifically, generative adversarial networks (GANs)~\cite{GAN} often yield sharper and more plausible textures than classical feed-forward encoders and witness success in image super-resolution~\cite{SRGAN} and in-painting~\cite{Inpainting}. Recently, \cite{kupyn2018deblurgan} introduced GAN to deblurring by treating it as a special image-to-image translation task~\cite{pix2pix}. The proposed model, called \textit{DeblurGAN}, was demonstrated to restore perceptually pleasing and sharp images, from both synthetic and real-world blurry images. DeblurGAN was also 5 times faster than its closest competitor as of then \cite{Nah2016DeepDeblurring}. 

Built on the success of DeblurGAN, this paper aims to make another substantial push on GAN-based motion deblurring. We introduce a new framework to improve over DeblurGAN, called \textbf{DeblurGAN-v2} in terms of both deblurring performance and inference efficiency, as well as to enable high flexibility over the quality- efficiency spectrum. Our innovations are summarized as below\footnote{An informal note: we quite like the sense of humor in \cite{redmon2018yolov3}, quoted as: "We present some updates to YOLO. We made a bunch of little design changes to make it better. We also trained this new network that's pretty swell." -- that well describes what we have done to DeblurGAN, too; although we consider DeblurGAN-v2 a non-incremental upgrade of DeblurGAN, with significant performance \& efficiency improvements.}:
\begin{itemize}
\vspace{-0.5em}
    \item \textbf{Framework Level:} We construct a new conditional GAN framework for deblurring. For the generator, we introduce the Feature Pyramid Network (FPN), which was originally developed for object detection \cite{lin2017feature}, to the image restoration task for the first time. For the discriminator, we adopt a relativistic discriminator \cite{jolicoeur2018relativistic} with a least-square loss wrapped \cite{LSGAN} inside, and with two columns that evaluate both global (image) and local (patch) scales respectively. 
    \vspace{-0.5em}
    \item \textbf{Backbone Level:} While the above framework is agnostic to the generator backbones, the choice would affect deblurring quality and efficiency. To pursue the state-of-the-art deblurring quality, we plug in a sophisticated Inception-ResNet-v2 backbone. To shift towards being more efficient, we adopt MobileNet, and further create its variant with depth-wise separable convolutions (MobileNet-DSC). The latter two become extremely compact in size and fast at inference.\vspace{-0.5em}
    \item \textbf{Experiment Level:} We present very extensive experiments on three popular benchmarks to show the state-of-the-art (or close) performance (PSNR, SSIM, and perceptual quality) achieved by DeblurGAN-v2. In terms of the efficiency, DeblurGAN-v2 with MobileNet-DSC is \textbf{11 times} faster than DeblurGAN \cite{kupyn2018deblurgan}, over \textbf{100 times} faster than \cite{Nah2016DeepDeblurring,tao2018scale}, and has a model size of just \textbf{4 MB}, implying the possibility of real-time video deblurring. We also present a subjective study of the deblurring quality on real blurry images. Lastly, we show the potential of our models in general image restoration, as extra flexibility.
   \vspace{-0.5em}
\end{itemize}

\section{Related work}
\label{s:related-work}
\subsection{Image Deblurring}
\vspace{-0.5em}

\begin{figure*}[htb]
  \includegraphics[width=0.99\textwidth]{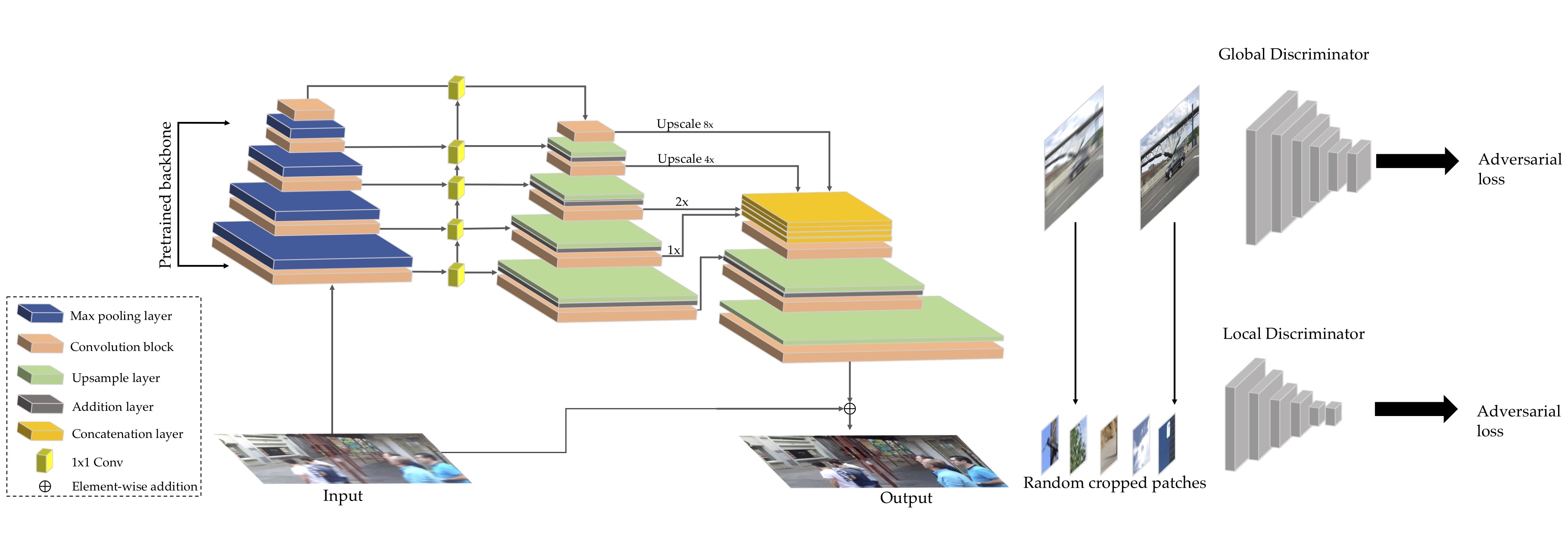}
  \vspace{-1.2em}
  \caption{DeblurGAN-v2 pipeline architecture.}
   \vspace{-1em}
  \label{fig:arch}
\end{figure*}

Single image motion deblurring is traditionally treated as a deconvolution problem, and can be tackled in either a blind or a non-blind manner. The former assumes a given or pre-estimated blur kernel \cite{ren2018deep,xu2018motion}. The latter is more realistic yet highly ill-posed. Earlier models rely on natural image priors to regularize deblurring \cite{normalized_sparsity,darkchannelprior,discriminativeprior,hyperlaplace}. However, most handcrafted priors cannot well capture the complicated blur variations in real images. 




Emerging deep learning techniques have boosted the breakthrough in image restoration tasks. Sun~\etal~\cite{SunLearningRemoval} exploited a convolutional neural network (CNN) for blur kernel estimation. Gong~\etal~\cite{GongFromBlur} used a fully convolutional network to estimate the motion flow. Besides those kernel-based methods, end-to-end kernel-free CNN methods were explored to restore a clean image from the blurry input directly, e.g., ~\cite{Nah2016DeepDeblurring, Noroozi2017MotionWild}. The latest work by Tao~\etal\cite{tao2018scale} extended the Multi-Scale CNN from \cite{Nah2016DeepDeblurring} to a Scale-Recurrent CNN for blind image deblurring, with impressive results.


The success of GANs for image restoration has impacted single image deblurring as well since Ramakrishnan~\etal~\cite{DeepGF} first solved image deblurring by referring to the image translation idea \cite{pix2pix}. Lately, Kupyn~\etal~\cite{kupyn2018deblurgan} introduced \textit{DeblurGAN} that exploited Wasserstein GAN~\cite{WGAN} with the gradient penalty~\cite{WGAN-GP} and the perceptual loss~\cite{Johnson2016Perceptual}.

\subsection{Generative adversarial networks}
\vspace{-0.5em}
A GAN \cite{GAN} consists of two
models: a discriminator $D$ and a generator $G$, that form a two-player minimax game. 
The generator learns to produce artificial samples and is trained to fool the discriminator, in a goal to capture the real data distribution. In particular, as a popular GAN variant, conditional GANs \cite{CGAN} have been widely applied to image-to-image translation problems, with image restoration and enhancement as special cases. They take the label or an observed image in addition to the latent code as inputs.


The minimax game with the value function $V(D, G)$ is formulated as the following \cite{GAN} (fake-real labels set to $0-1$):
\begin{align}\nonumber
\vspace{-0.5em}
\min_G \max_D V(D,G) & = \mathbb{E}_{x\sim p_{data}(x)} \big[\log D(x) \big] \\ & + \mathbb{E}_{z\sim p_{z}(z)} \big[\log (1 - D(G(z))) \big] \vspace{-0.5em}
\nonumber
\end{align}
Such an objective function is notoriously hard to optimize, and one needs to deal with many challenges, e.g., mode collapse and gradient vanishing/explosion, during the training process. To fix the vanishing gradients and stabilize the training, Least Squares GANs discriminator \cite{LSGAN} tried to introduce a loss function that provides smoother and non-saturating gradient. 
The authors observe that the log-type loss in \cite{GAN} saturates quickly as it ignores the distance between $x$ to the decision boundary. In contrast, an $L2$ loss provides gradients proportional to that distance, so that fake samples more far away from the boundary receive larger penalties. 
The proposed loss function also minimizes the Pearson $\chi^2$ divergence that leads to the better training stability. 
The LSGAN objective function is written as::
\begin{align}\nonumber
\min_D V(D) & = \frac{1}{2} \mathbb{E}_{x\sim p_{data}(x)} \big[(D(x) - 1)^2 \big] \\ & +
\frac{1}{2} \mathbb{E}_{z\sim p_{z}(z)} \big[D(G(z))^2 \big] \\
\min_G V(G) & = \frac{1}{2} \mathbb{E}_{z\sim p_{z}(z)} \big[(D(G(z))-1)^2 \big] \nonumber
\end{align}




Another relevant improvement to GANs is the Relativistic GAN \cite{jolicoeur2018relativistic}. It used a relativistic discriminator to estimate the probability that the given real data is more realistic than a randomly sampled fake data. As the author advocated, such would account for a priori knowledge that half of the data in the mini-batch is fake. 
The relativistic discriminators show more stable and computationally efficient training in comparison to other GAN types, including WGAN-GP \cite{WGAN-GP} that was used in DeblurGAN-v1.


\vspace{-0.2em}
\section{DeblurGAN-v2 Architecture}
\vspace{-0.5em}
The overview of DeblurGAN-v2 architecture is illustrated in Figure \ref{fig:arch}. It restores a sharp image $I_S$ from a single blurred image $I_B$, via the trained generator.


\vspace{-0.2em}
\subsection{Feature Pyramid Deblurring}
\vspace{-0.4em}
Existing CNNs for image deblurring (and other restoration problems)~\cite{SRGAN,Nah2016DeepDeblurring} typically refer to ResNet-like structures. Most state-of-the-art methods \cite{Nah2016DeepDeblurring,tao2018scale} dealt with different levels of blurs, utilizing multi-stream CNN s with an input image pyramid at different scales. However, processing multiple scale images is time-consuming and memory-demanding. We introduce the idea of Feature Pyramid Networks \cite{lin2017feature} to image deblurring (more generally, the field of image restoration and enhancement), \textit{for the first time to our best knowledge}. We treat this novel approach as a lighter-weight alternative to incorporate multi-scale features. 

The FPN module was originally designed for object detection \cite{lin2017feature}. It generates multiple feature map layers which encode different semantics and contain better quality information. FPN comprises a bottom-up and a top-down pathway. The bottom-up pathway is the usual convolutional network for feature extraction, along which the spatial resolution is downsampled, but more semantic context information is extracted and compressed. Through the top-down pathway, FPNs reconstructs higher spatial resolution from the semantically rich layers. The lateral connections between the bottom-up and top-down pathways supplement high-resolution details and help localize objects. 

Our architecture consists of an FPN backbone from which we take five final feature maps of different scales as the output. Those features are later up-sampled to the same $\frac{1}{4}$ input size and concatenated into one tensor which contains the semantic information on different levels. We additionally add two upsampling and convolutional layers at the end of the network to restore the original image size  and reduce artifacts. Similar to \cite{kupyn2018deblurgan,liu2018image}, we introduce a direct skip connection from the input to the output, so that the learning focuses on the residue. The input images are normalized to [-1 1]. We also use a \textit{tanh} activation layer to keep the output in the same range. In addition to the multi-scale feature aggregation capability, FPN also strikes a balance between accuracy and speed: please see experiment parts.

\vspace{-0.2em}
\subsection{Choice of Backbones: Trade-off between Performance and Efficiency}
\vspace{-0.5em}
The new FPN-embeded architecture is agnostic to the choice of feature extractor backbones. With this plug-and-play property, we are entitled with the flexibility to navigate through the spectrum of accuracy and efficiency. By default, we choose ImageNet-pretrained backbones to convey more semantic-related features. As one option, we use \textbf{Inception-ResNet-v2} \cite{szegedy2017inception} to pursue strong deblurring performance, although we find other backbones such as SE-ResNeXt \cite{hu2018squeeze} to be similarly effective. 

The demands of efficient restoration model have recently drawn increasing attentions due to the prevailing need of mobile on-device image enhancement \cite{yu2018crafting,wu2018deep,wang2018energynet}. To explore this direction, we choose the \textbf{MobileNet} V2 backbone \cite{MobileNet} as one option. To reduce the complexity further, we try another more aggressive option on top of DeblurGAN-v2 with MobileNet V2, by replacing all normal convolutions in the \textit{full network} (including those not in backbone) with Depthwise Separable Convolutions \cite{chollet2017xception}. The resulting model is denoted as \textbf{MobileNet-DSC}, and can provide extremely lightweight and efficient image deblurring.

To unleash this important flexibility to practitioners, in our codes, we have implemented the switch of backbones as a simple \textit{one-line command}: it can be compatible with many state-of-the-art pre-trained networks. 


\subsection{Double-Scale RaGAN-LS Discriminator}
\vspace{-0.5em}
\label{s:method}

  
Instead of the WGAN-GP discriminator in DeblurGAN \cite{kupyn2018deblurgan}, we suggest several upgrades in DeblurGAN-v2. We first adopt the relativistic ``wrapping'' \cite{jolicoeur2018relativistic} on the LSGAN \cite{LSGAN} cost function, creating a new \textit{RaGAN-LS} loss:
\begin{align}\label{obj:NDFT_final}\nonumber
\vspace{-1em}
L_D^{RaLSGAN} & =  \mathbb{E}_{x\sim p_{data}(x)} \big[(D(x) -  \mathbb{E}_{z\sim p_{z}(z)} D(G(z)) - 1)^2 \big] \\ 
& + \mathbb{E}_{z\sim p_{z}(z)} \big[(D(G(z))- \mathbb{E}_{x\sim p_{data}(x)} D(x) + 1)^2 \big]
\vspace{-1em}
\end{align}
It is observed to make training notably faster and more stable compared to using the WGAN-GP objective. 
We also empirically conclude that the generated results possess higher perceptual quality and overall sharper outputs. Correspondingly, the adversarial loss $L_{adv}$ for the DeblurGAN-v2 generator will be optimizing (\ref{obj:NDFT_final}) w.r.t. $G$.




\textbf{Extending to Both Global and Local Scales}.
 Isola~\etal\cite{pix2pix} propose to use a PatchGAN discriminator which operates on the images patches of size 70 $\times$ 70, that proves to produce sharper results than the standard ``global'' discriminator that operates on the full image. The PatchGAN idea was adopted in DeblurGAN \cite{kupyn2018deblurgan}.
 
 However, we observed that for highly non-uniform blurred images, especially when complex object movements are involved, the ``global'' scales are still essential for discriminators to incorporate full spatial contexts \cite{jiang2019enlightengan}. To take advantage of both global and local features, we propose to use a double-scale discriminator, consisting of one local branch that operates on patch levels like \cite{pix2pix} did, and the other global branch that feeds the full input image. We observe that to allow DeblurGAN-v2 to better handle larger and more heterogeneous real blurs. 
 
 
\textbf{Overall Loss Function}
For training image restoration GANs, one needs to compare the images on the training stage – the reconstructed and the original ones, under some metric. One common option is the pixel-space loss $L_P$, e.g., the simplest $L_1$ or $L_2$ distance.  
  As \cite{SRGAN} suggested, using $L_p$ tends to yield oversmoothened pixel-space outputs. 
  \cite{kupyn2018deblurgan} proposed to use the perceptual distance \cite{Johnson2016Perceptual}, as a form of ``content'' loss $L_X$. In contrast to the $L_2$, it computes the Euclidean loss on the VGG19 \cite{VGGNet2014} \textit{conv3\_3} feature maps.  We incorporate those prior wisdoms and use a hybrid three-term loss for training DeblurGAN-v2:
$$L_G = 0.5 * L_p + 0.006 * L_{X} + 0.01 * L_{adv}$$
The $L_{adv}$ terms contains both global and local discriminator losses. Also, we choose mean-square-error (MSE) loss as $L_p$: although DeblurGAN did not include an $L_p$ term, we find it to help correct color and texture distortions.
  


  
\subsection{Training Datasets}
\vspace{-0.5em}
\begin{figure}[htp!]
\centering
\includegraphics[width=0.23\textwidth]{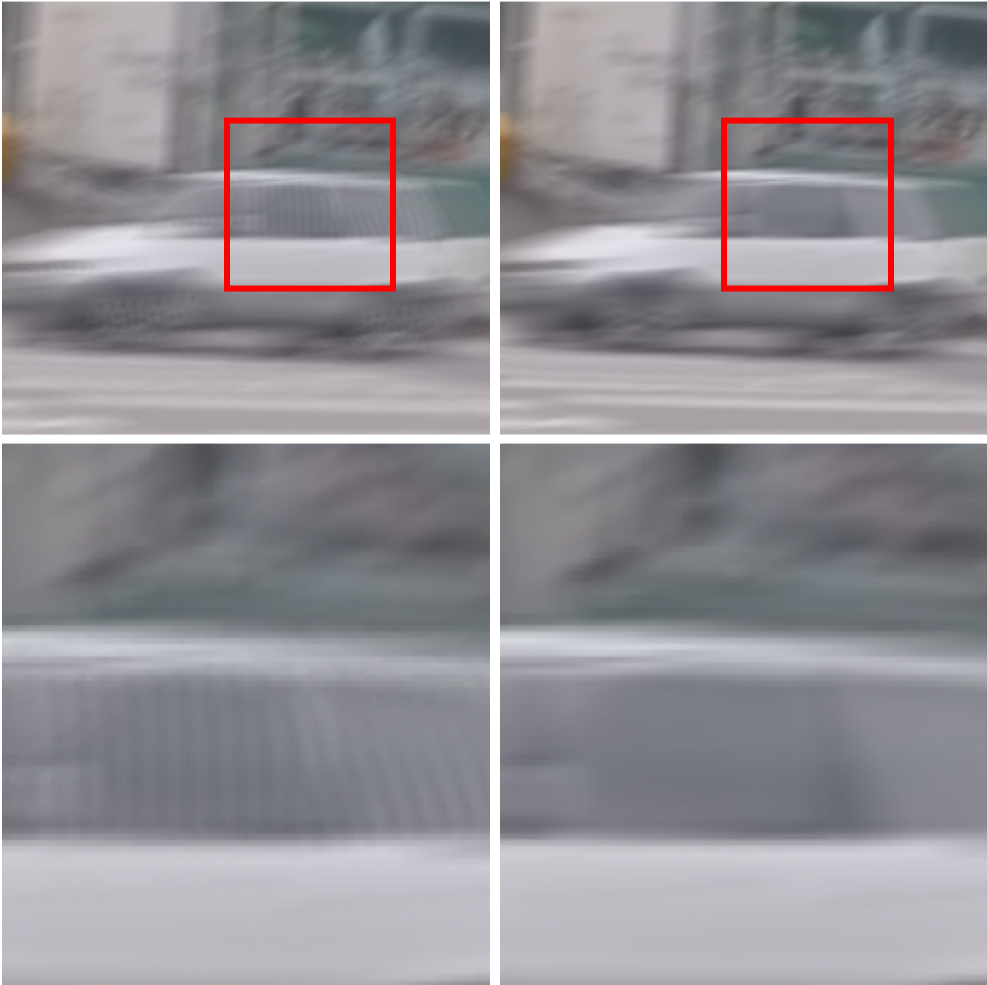}
\includegraphics[width=0.23\textwidth]{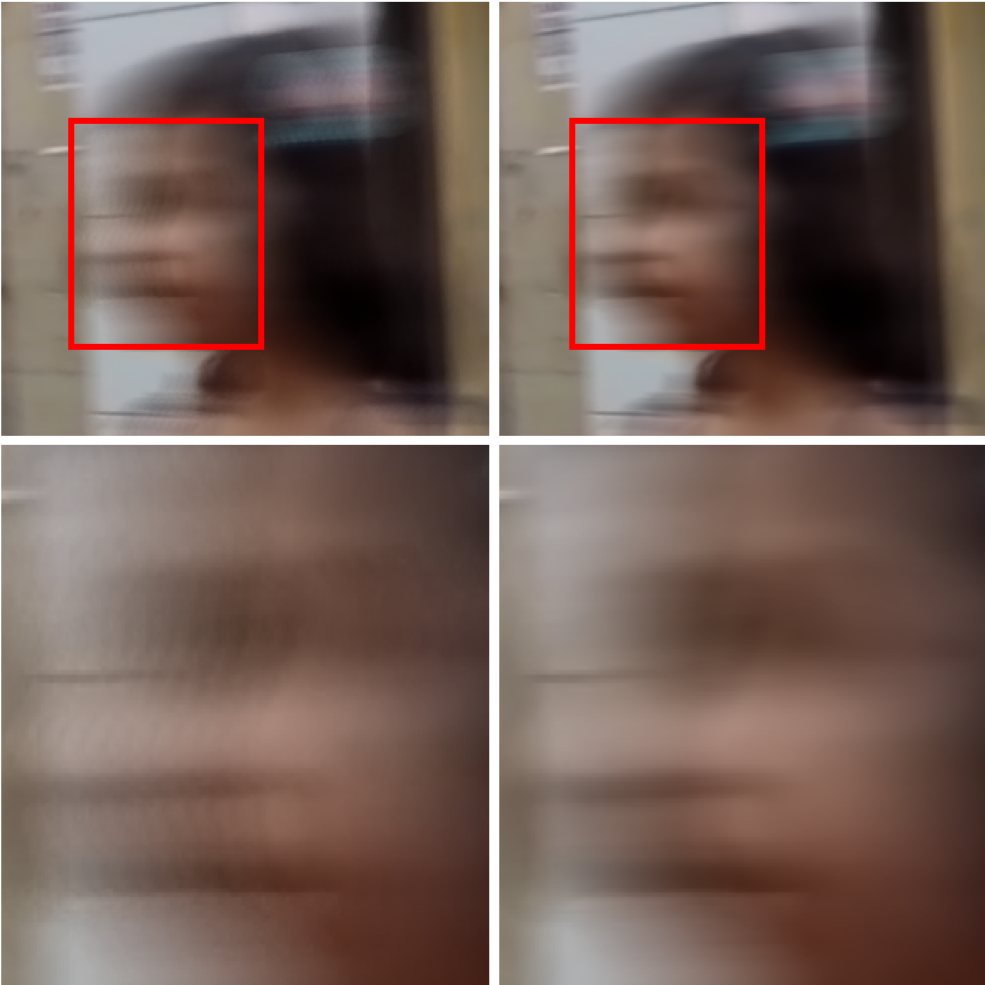}
\\ \small{(a) \qquad\qquad \quad(b) \qquad \quad  \qquad (c)\qquad\qquad \quad (d)}
\vspace{-0.5em}
\caption{Visual comparison of synthesized blurry images, without interpolation (a,c) and with interpolation (b,d).}
\label{fig:interp_compare}
\vspace{-1em}
\end{figure}

\begin{table*}[htb]
\small
\caption{Performance and efficiency comparison on the GoPro test dataset, All models were tested on the \emph{linear} image subset.}
\vspace{-0.5em}
\label{T:gopro}
\centering
\ra{1.0}
\setlength\tabcolsep{3pt}
\begin{tabular}{cccccc|ccc}
\toprule
 &  Sun~\etal~\cite{SunLearningRemoval} & Xu~\etal~\cite{XuUnnaturalDeblurring} & DeepDeblur~\cite{Nah2016DeepDeblurring} & SRN~\cite{tao2018scale} & DeblurGAN~\cite{kupyn2018deblurgan} & Inception-ResNet-v2  & MobileNet & MobileNet-DSC \\
\hline
PSNR & 24.64  & 25.10 & 29.23& \textbf{30.10} & 28.70 & 29.55  & 28.17 & 28.03 \\
SSIM & 0.842  &0.890 & 0.916 & 0.932 & 0.927 & \textbf{0.934}  & 0.925 & 0.922 \\
\midrule
Time & 20 min & 13.41s & 4.33s& 1.6s & 0.85s & 0.35s  & 0.06s & \textbf{0.04s} \\
FLOPS & N/A & N/A & 1760.04G & 1434.82G & 678.29G & 411.34G & 43.75G & \textbf{14.83G} \\
\midrule
\bottomrule
\end{tabular}
\vspace{-0.5em}
\end{table*}

\begin{table*}[htb]
\small
\caption{PSNR and SSIM comparison on the Kohler dataset. 
}
\vspace{-0.5em}
\label{T:kohler}
\centering
\begin{tabular}{ccccc|ccc}
\toprule
Method & Sun~\cite{SunLearningRemoval} & DeepDeblur~\cite{Nah2016DeepDeblurring} & SRN \cite{tao2018scale} & DeblurGAN~\cite{kupyn2018deblurgan} & Inception-ResNet-v2 & MobileNet  & MobileNet-DSC\\ 
\hline
PSNR & 25.22 & 26.48 & 26.75  & 26.10 & 26.72 & 26.36 & 26.35 \\
SSIM & 0.773  & 0.807 & 0.837  & 0.816 & 0.836 & 0.820 & 0.819  \\
\bottomrule
\end{tabular}
\vspace{-1em}
\end{table*}

The \textbf{GoPro} dataset \cite{Nah2016DeepDeblurring} uses the GoPro Hero 4 camera to capture 240 frames per second (fps) video sequences, and generate blurred images through averaging consecutive short-exposure frames. It is a common benchmark for image motion blurring, containing 3,214 blurry/clear image pairs. We follow the same split \cite{Nah2016DeepDeblurring}, to use 2,103 pairs for training and the remaining 1,111 pairs for evaluation.


The \textbf{DVD} dataset \cite{su2017deep} collects 71 real-world videos captured by various devices such as iPhone 6s, GoPro Hero 4 and Nexus 5x, at 240 fps. The author then generated 6708 synthetic blurry and sharp pairs by averaging consecutive short-exposure frames to approximate a longer exposure \cite{telleen2007synthetic}. The dataset was initially used for video deblurring but was later also brought to the image deblurring field.


The \textbf{NFS} dataset \cite{kiani2017need} was initially proposed to benchmark visual object tracking. It consists of 75 videos captured with high-frame rate cameras from iPhone 6 and iPad Pro. Additionally, 25 sequences are collected from YouTube captured at 240
fps from a variety of different devices. It covers variety of scenes including sport, skydiving, underwater, wildlife, roadside, and indoor scenes.

\textbf{Training data preparation:} Conventionally, the blurry frames are averaged from consecutive clean frames. However, we notice unrealistic ghost effects when observing the directly averaged frames, as in Figure \ref{fig:interp_compare}(a)(c). To alleviate that, we first use a video frame interpolation model \cite{niklaus2017video} to increase the original 240-fps videos to 3840 fps, then perform average pooling over the same time window (but now with more frames). It leads to smoother and more continuous blurs, as in Figure \ref{fig:interp_compare}(b)(d). Experimentally, this data preparation did not noticeably impact PSNR/SSIM but was observed to improve the visual quality results.












\vspace{-0.2em}
\section{Experimental evaluation}
\vspace{-0.4em}

\begin{figure}[htb]
	\centering
    \captionsetup{justification=centering}
        \includegraphics[width=0.225\textwidth]{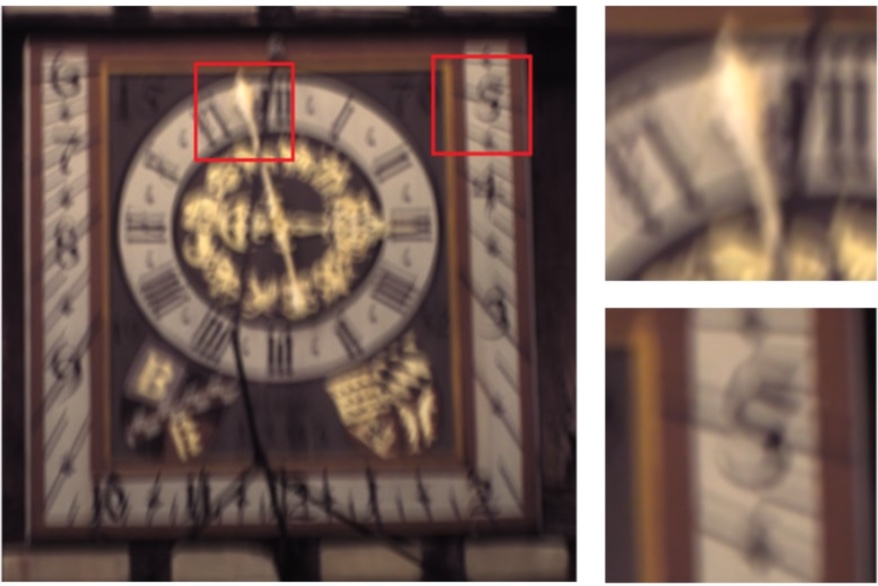}
        \includegraphics[width=0.225\textwidth]{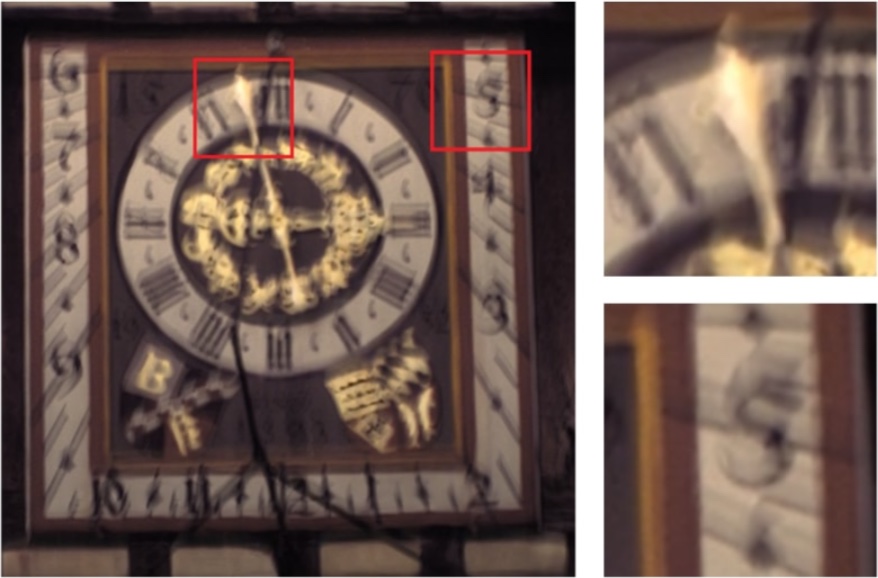}
    \vspace{0.25em}
        \includegraphics[width=0.225\textwidth]{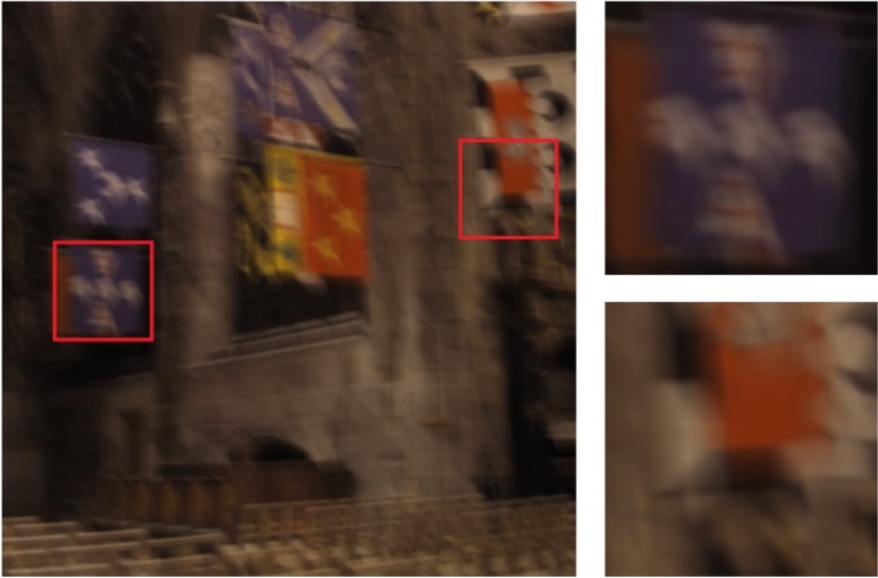}
        \includegraphics[width=0.225\textwidth]{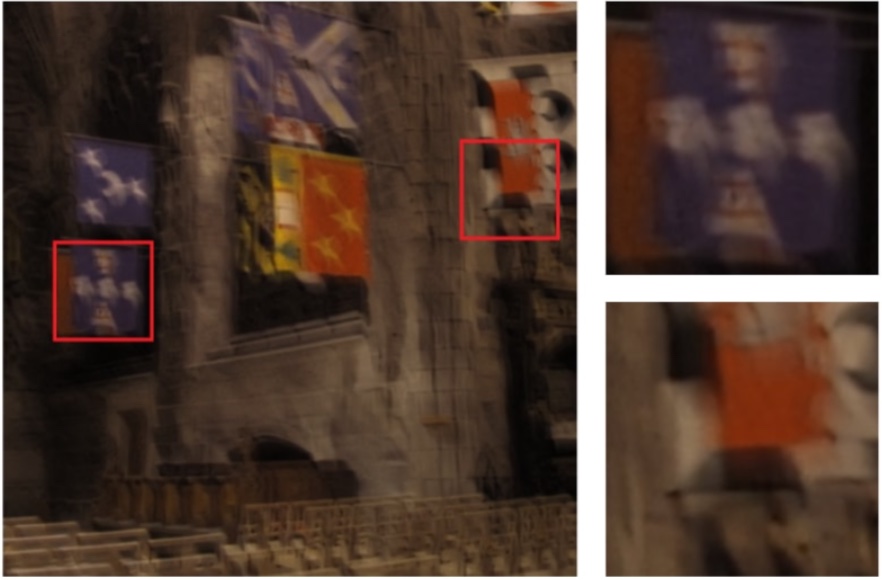}
          \\  \small{(a) Blurry \qquad \qquad \qquad (b) DeepDeblur \cite{Nah2016DeepDeblurring}}
          \\
          \vspace{0.25em}
        \includegraphics[width=0.225\textwidth]{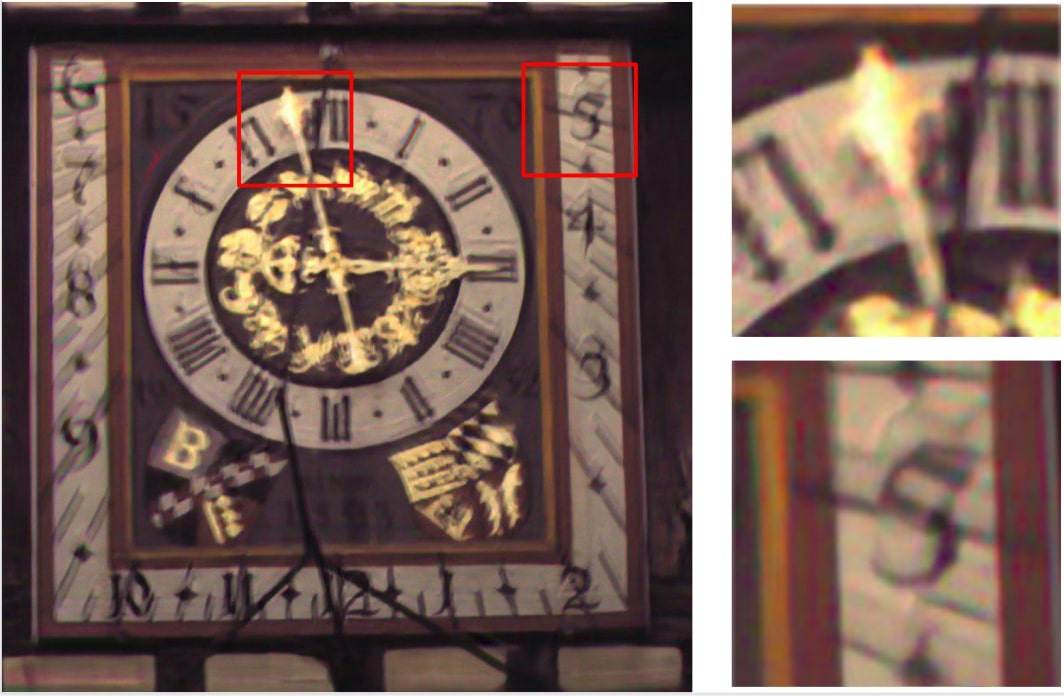}
        \includegraphics[width=0.225\textwidth]{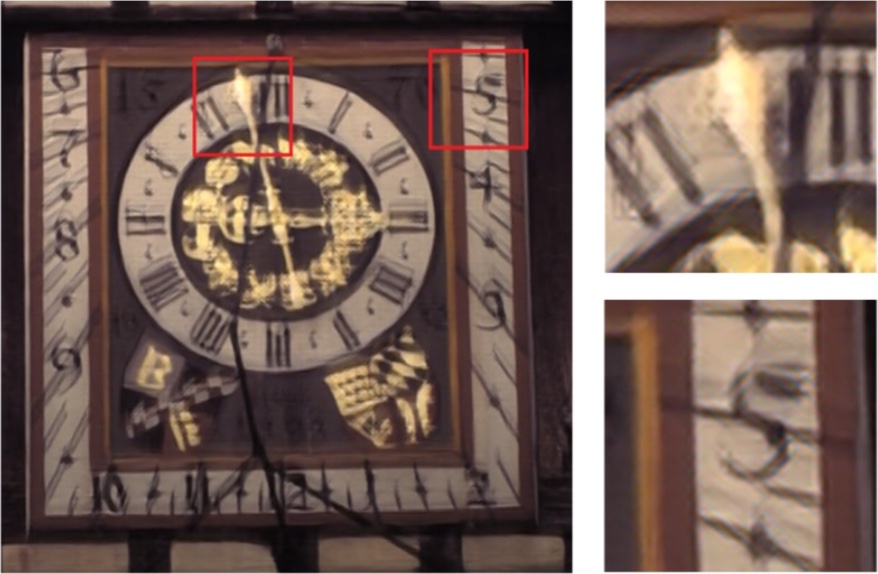}
    \vspace{0.25em}
        \includegraphics[width=0.225\textwidth]{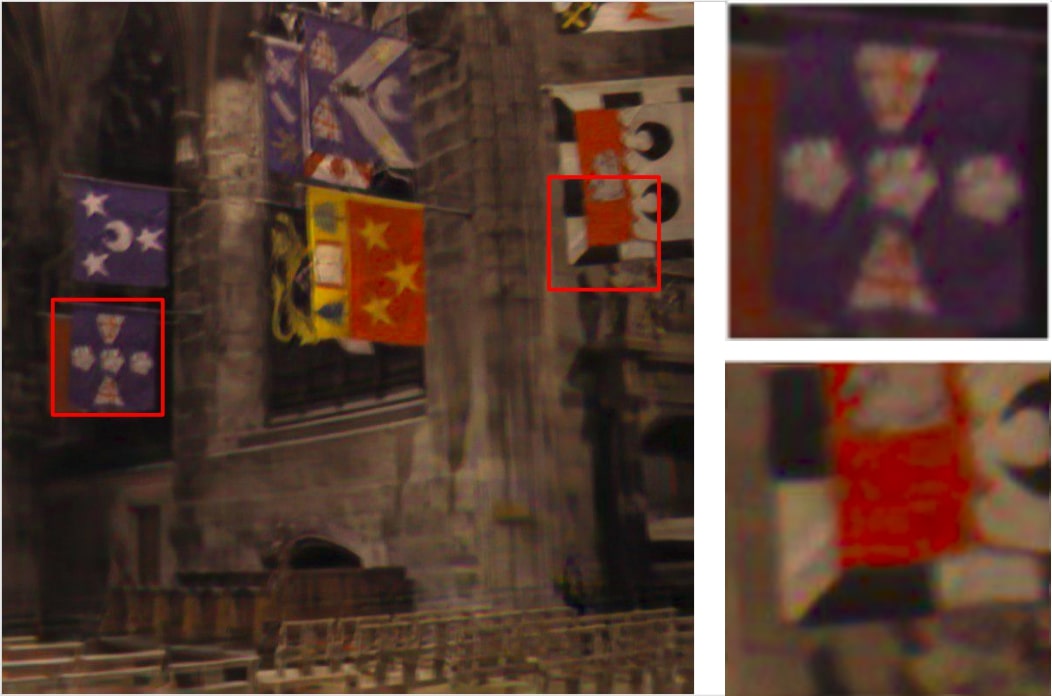}
        \includegraphics[width=0.225\textwidth]{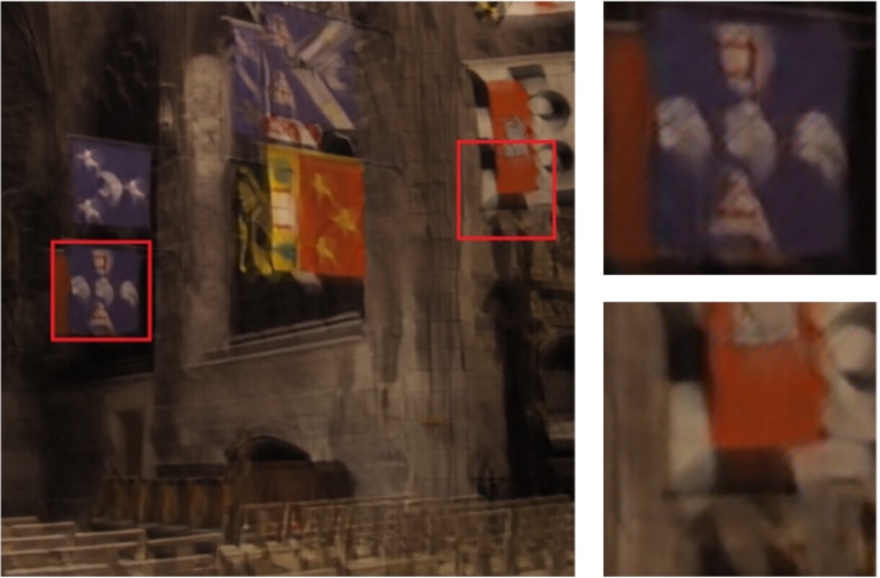}
        \\  \small{(c) SRN \cite{tao2018scale} \qquad \qquad \qquad (d) DeblurGAN \cite{kupyn2018deblurgan}}
          \\
    \vspace{0.25em}
        \includegraphics[width=0.225\textwidth]{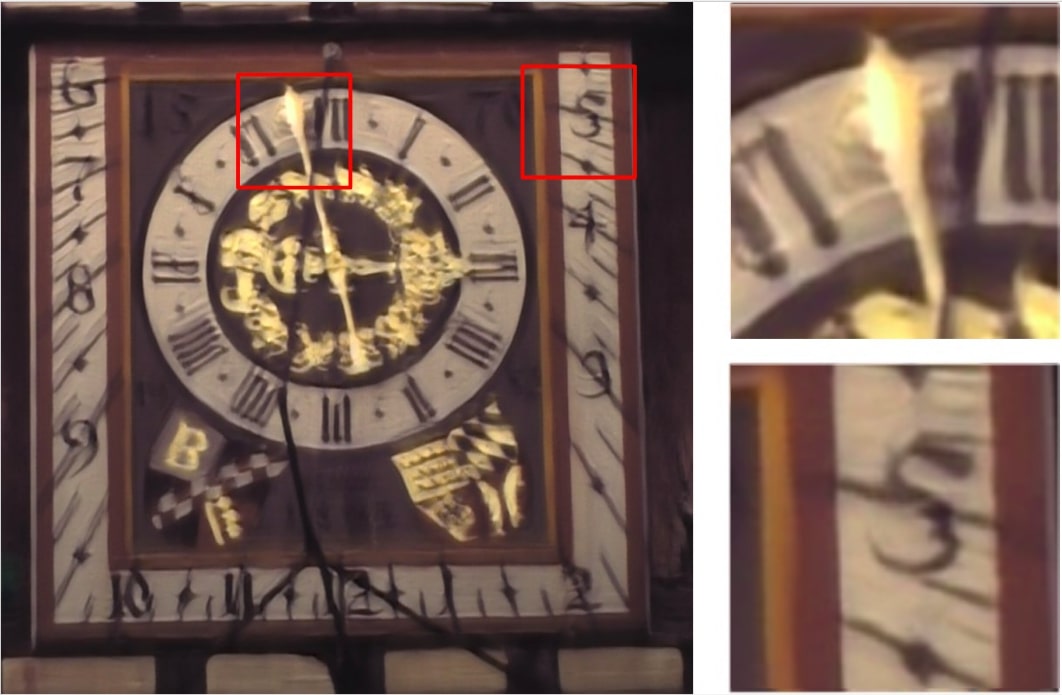}
        \includegraphics[width=0.225\textwidth]{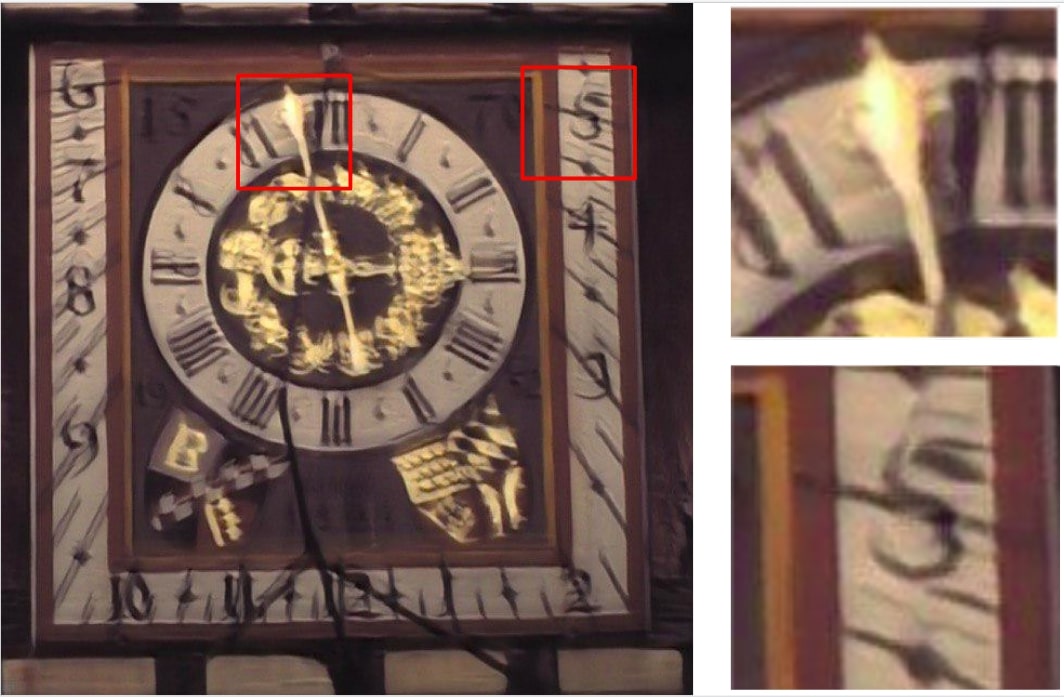}
    \vspace{0.25em}
        \includegraphics[width=0.225\textwidth]{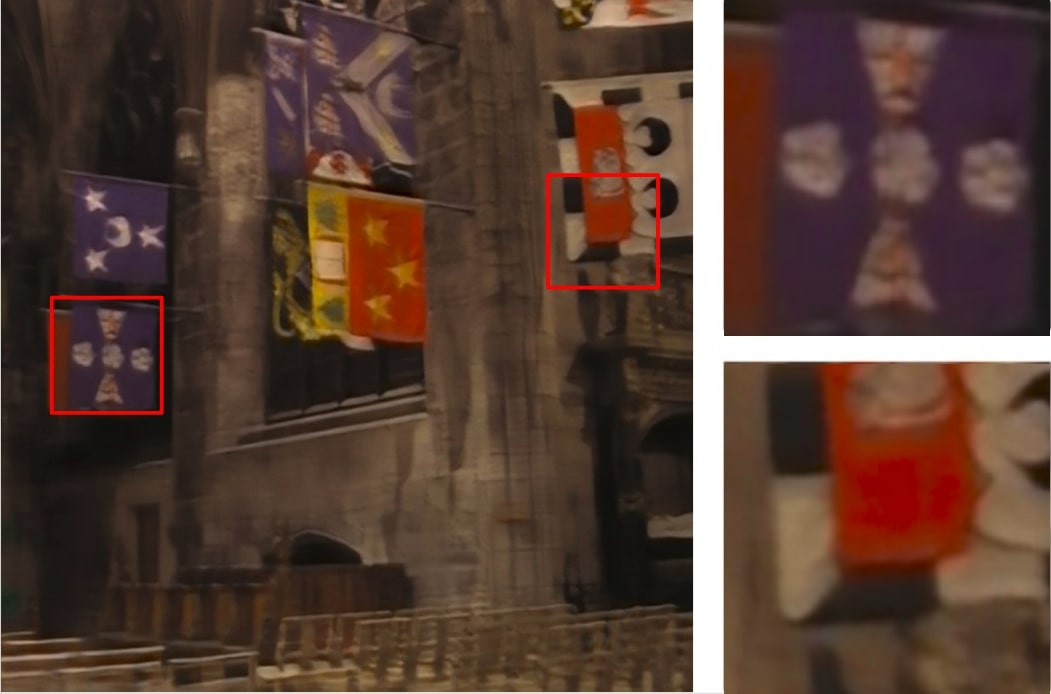}
        \includegraphics[width=0.225\textwidth]{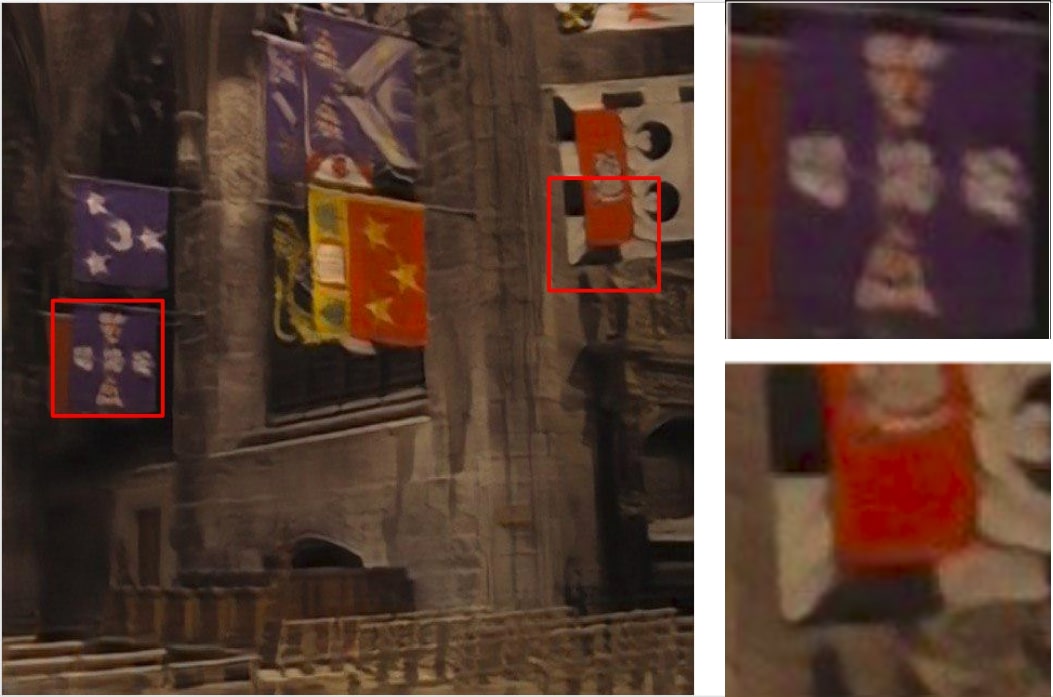}
        \\\footnotesize{(e) DeblurGAN-v2 \qquad \qquad \qquad (f) DeblurGAN-v2 \\ \quad (Inception-ResNet-v2)  \qquad \qquad \qquad \quad (MobileNet) \qquad}
            \vspace{-0.5em}
    \caption{Visual comparison on the Kohler dataset.}
    \vspace{-1em}
    \label{fig:kohler_test}
\end{figure}

\subsection{Implementation Details}
\vspace{-0.5em}
We implemented all of our models using PyTorch~\cite{pytorch}. We compose our training set by selecting each second frame from the GoPro and DVD datasets, and every tenth frame from the NFS dataset, with the hope to reduce overfitting to any specific dataset. We then train DeblurGAN-v2 on the resulting set of approximately 10,000 image pairs. Three backbones are evaluated: Inception-ResNet-v2, MobileNet, and MobileNet-DSC. The former targets at high-performance deblurring, while the latter two are more suited for resource-constrained edge applications. Specifically, the extremely lightweight DeblurGAN-v2 (MobileNet-DSC) costs 96\% fewer parameters than DeblurGAN-v2 (Inception-ResNet-v2). 

All models were trained on a single Tesla-P100 GPU, with Adam~\cite{ADAM} optimizer and the learning rate of $10^{-4}$ for 150 epochs, followed by another 150 epochs with a linear decay to $10^{-7}$. We freeze the pre-trained backbone weights for 3 epochs, and then we unfreeze all weights and continue the training. The un-pre-trained parts are initialized with random Gaussian. The training takes 5 days to converge. The models are fully convolutional, thus can be applied to the images of arbitrary size. 

\label{s:results}
\begin{figure*}[htb]
	\centering
    \begin{subfigure}[t]{0.225\textwidth}
        \includegraphics[width=\textwidth]{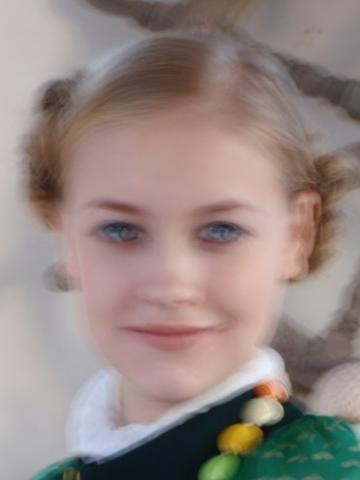}
        \caption{Blurred photo}
        \label{fig:lai_in}
    \end{subfigure}
    \begin{subfigure}[t]{0.225\textwidth}
        \includegraphics[width=\textwidth]{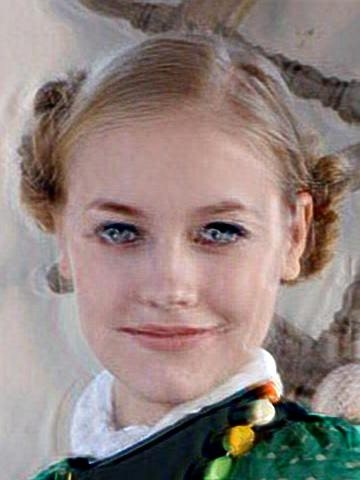}
        \caption{Whyte~\etal~\cite{whyte}}
        \label{fig:lai_whyte}
    \end{subfigure}
    \begin{subfigure}[t]{0.225\textwidth}
        \includegraphics[width=\textwidth]{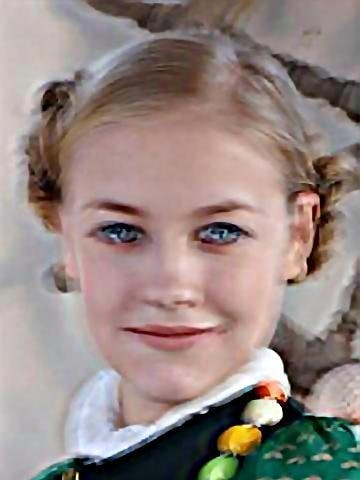}
        \caption{Krishnan~\etal \cite{normalized_sparsity}}
        \label{fig:lai_k}
    \end{subfigure}
    \begin{subfigure}[t]{0.225\textwidth}
        \includegraphics[width=\textwidth]{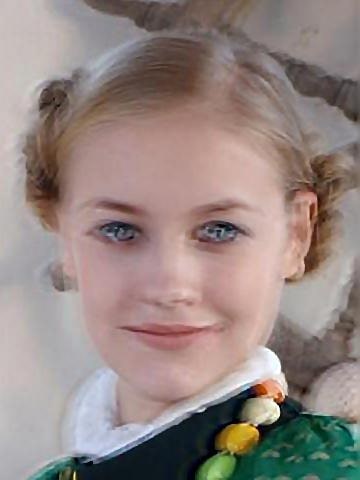}
        \caption{Sun~\etal~\cite{SunLearningRemoval}}
        \label{fig:lai_sun}
    \end{subfigure}

    \bigskip
	\centering
    \begin{subfigure}[t]{0.225\textwidth}
        \includegraphics[width=\textwidth]{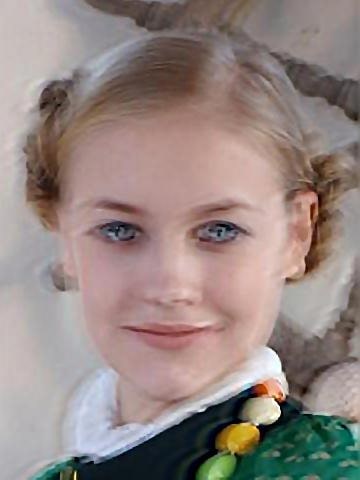}
        \caption{Xu~\etal ~\cite{XuUnnaturalDeblurring}}
        \label{fig:lai_xu}
    \end{subfigure}
    \begin{subfigure}[t]{0.225\textwidth}
        \includegraphics[width=\textwidth]{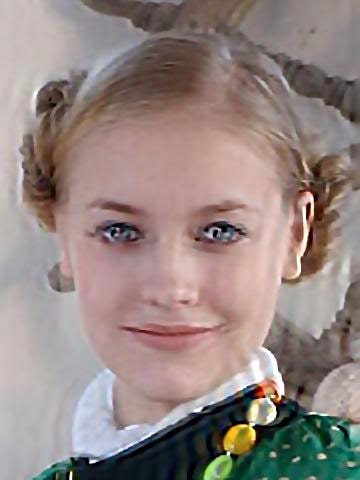}
        \caption{Pan~\etal~\cite{darkchannelprior}}
        \label{fig:lai_whyte}
    \end{subfigure}
        \begin{subfigure}[t]{0.225\textwidth}
        \includegraphics[width=\textwidth]{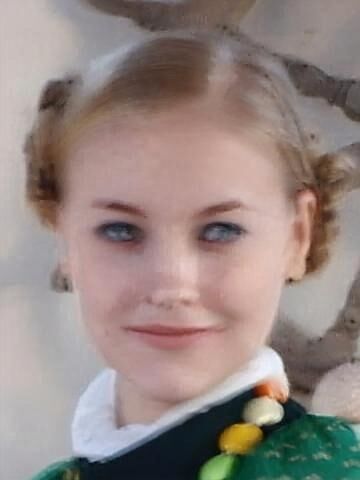}
        \caption{DeepDeblur \cite{Nah2016DeepDeblurring}}
        \label{fig:lai_dd}
    \end{subfigure}
    \begin{subfigure}[t]{0.225\textwidth}
        \includegraphics[width=\textwidth]{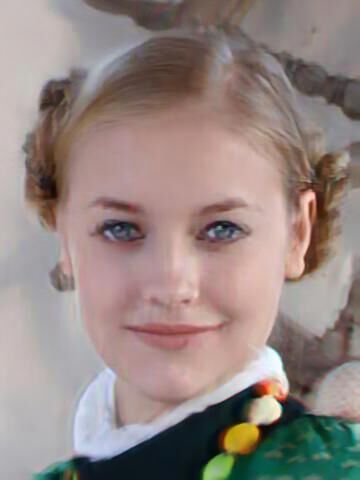}
        \caption{SRN \cite{tao2018scale}}
        \label{fig:lai_k}
    \end{subfigure}

    \bigskip
	\centering
        \begin{subfigure}[t]{0.225\textwidth}
        \includegraphics[width=\textwidth]{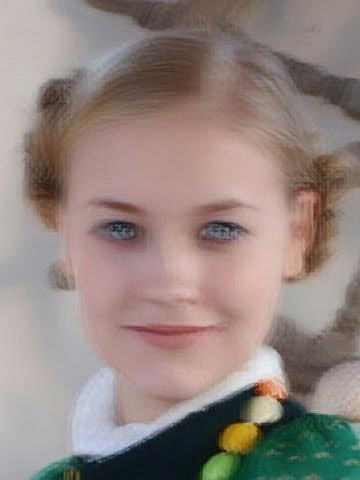}
        \caption{DeblurGAN \cite{kupyn2018deblurgan}}
        \label{fig:lai_dg}
    \end{subfigure}
    \begin{subfigure}[t]{0.225\textwidth}
        \includegraphics[width=\textwidth]{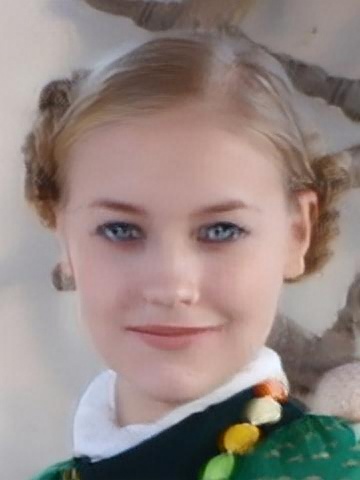}
        \caption{DeblurGAN-v2 \\ (Inception-ResNet-v2)} \footnotesize{[\textbf{Best visual quality}]}
        \label{fig:lai_our}
    \end{subfigure}
    \begin{subfigure}[t]{0.225\textwidth}
        \includegraphics[width=\textwidth]{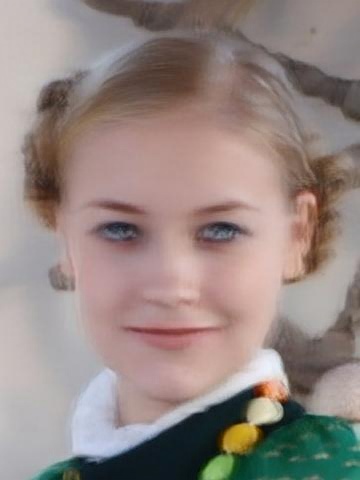}
        \caption{DeblurGAN-v2 \\ (MobileNet)} \footnotesize{[\textbf{High efficiency}]}
        \label{fig:lai_sharp}
    \end{subfigure}
    \begin{subfigure}[t]{0.225\textwidth}
        \includegraphics[width=\textwidth]{face2_mobilenet}
        \caption{DeblurGAN-v2 \\(MobileNet-DSC)} \footnotesize{[\textbf{Highest efficiency]}}
        \label{fig:lai_sharp2}
    \end{subfigure}
    \vspace{-0.5em}
    \caption{Qualitative comparison on the ``face2'' test image of the Lai dataset \cite{lai2016comparative}. DeblurGAN-v2 models are artifact-free, in contrast to other neural and non-CNN algorithms, producing smoother and visually more pleasing results.}\label{fig:lai}
    \vspace{-1em}
\end{figure*}

 \begin{figure*}[htb]
 \vspace{-1em}
	\centering
    \captionsetup{justification=centering}
    \begin{subfigure}[t]{0.24\textwidth}
        \includegraphics[width=\textwidth]{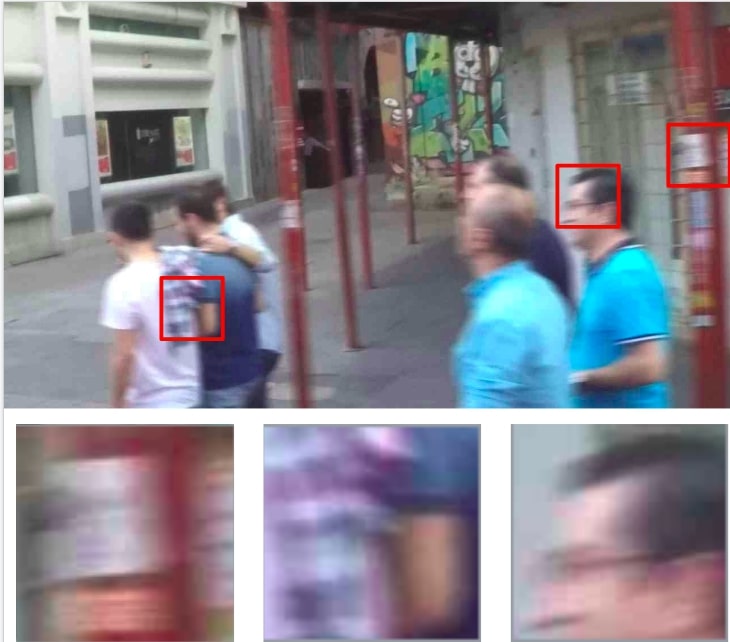}
        \caption{Degraded photo}
        \label{fig:rest_input}
    \end{subfigure}
    \begin{subfigure}[t]{0.24\textwidth}
        \includegraphics[width=\textwidth]{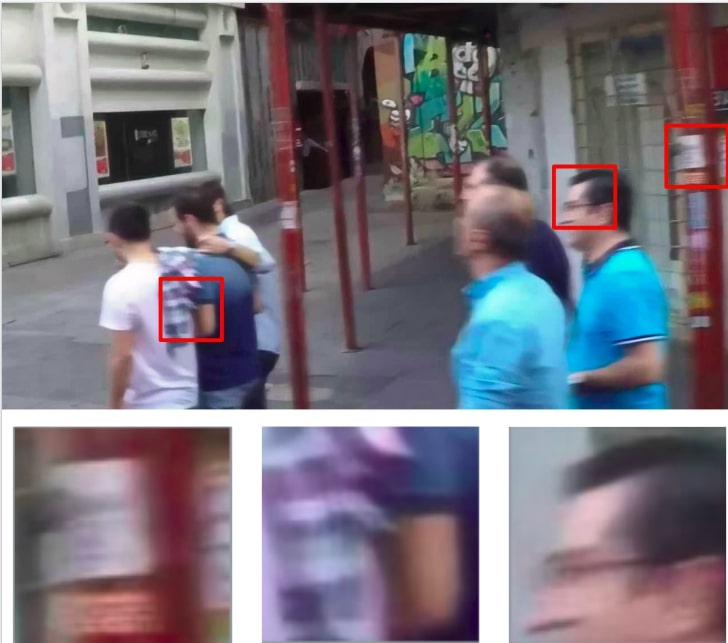}
        \caption{DeblurGAN}
        \label{fig:rest_dg}
    \end{subfigure}
    \begin{subfigure}[t]{0.24\textwidth}
        \includegraphics[width=\textwidth]{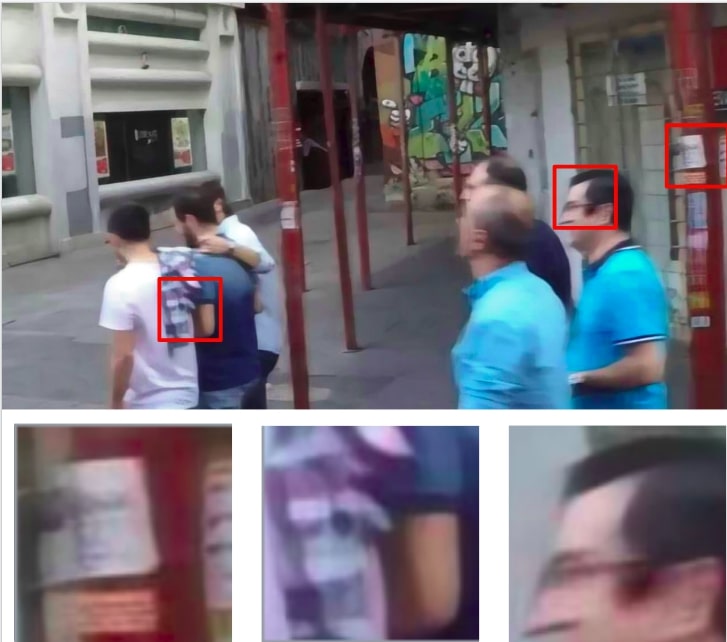}
        \caption{DeblurGAN-v2 \footnotesize{(Inception-ResNet-v2)}}
        \label{fig:rest_our}
    \end{subfigure}
    \begin{subfigure}[t]{0.24\textwidth}
        \includegraphics[width=\textwidth]{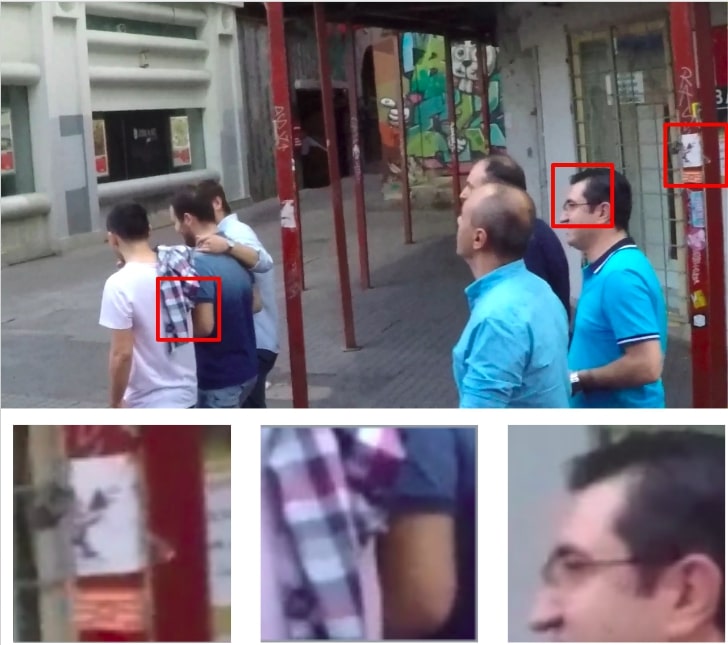}
        \caption{Clean photo}
        \label{fig:rest_gt}
    \end{subfigure}
    \vspace{-1em}
    \caption{Visual comparison example on the Restore Dataset.}
        \vspace{-1.2em}
    \label{fig:yolo}
\end{figure*}

\subsection{Quantitative Evaluation on GoPro Dataset}
\vspace{-0.5em}


We compare our models with a number of state-of-the-arts: one of is a traditional method by Xu \etal \cite{XuUnnaturalDeblurring}, while the rest are deep learning-based: ~\cite{SunLearningRemoval} by Sun \etal, DeepDeblur \cite{Nah2016DeepDeblurring}, SRN \cite{tao2018scale}, and DeblurGAN \cite{kupyn2018deblurgan}. We compare on both standard performance metrics (PSNR, SSIM), and inference efficiency (averaged running time per image measured on a single GPU). Results are summarized in Table\ref{T:gopro}. 

In terms of PSNR/SSIM, DeblurGAN-v2 (Inception-ResNet-v2) and SRN are ranked top-2: DeblurGAN-v2 (Inception-ResNet-v2) has slightly lower PSNR, which is not surprising since it was not trained under pure MSE loss; but it outperforms SRN in SSIM. However, we are very encouraged to observe that DeblurGAN-v2 (Inception-ResNet-v2) takes \textbf{78\% less} inference time than SRN. Moreover, two of our light-weight models, DeblurGAN-v2 (MobileNet) and DeblurGAN-v2 (MobileNet-DSC), show SSIMs (0.925 and 0.922) on par with the other two latest deep deblurring methods, DeblurGAN (0.927) and DeepDeblur (0.916), while being up to \textbf{100 times faster}. 


In particular, MobileNet-DSC only costs 0.04s per image, which even enables near real-time video frame deblurring, for 25-fps videos. To our best knowledge, DeblurGAN-v2 (MobileNet-DSC) is the only deblurring method so far that can simultaneously achieve (reasonably) high performance and that high inference efficiency.




\subsection{Quantitative Evaluation on Kohler dataset}
\vspace{-0.5em}
The Kohler dataset ~\cite{Kohler} consists of 4 images, each blurred with 12 different kernels. It is a standard benchmark for evaluating blind deblurring algorithms. The dataset was generated by recording and analyzing real camera motion, which was then played back on a robot platform such that a sequence of sharp images was recorded sampling the 6D camera motion trajectory.

\begin{table}[tb]
\small
\caption{Results on DVD dataset}
\label{T:restore}
\vspace{-0.5em}
\centering
\ra{1.0}
\setlength\tabcolsep{3pt}
\begin{tabular}{ccccc}
\toprule
 & PSNR & SSIM & Inference Time & Resolution\\
 \hline
WFA & 28.35 &  N/A  & N/A & N/A \\
DVD (single) & 28.37 &  0.913  & 1.0s & 960 x 540 \\
DeblurGAN-v2 & \multirow{2}{*}{\textbf{28.54}} &  \multirow{2}{*}{\textbf{0.929}}  & \multirow{2}{*}{\textbf{0.06s}} & \multirow{2}{*}{1280 x 720}\\
(MobileNet)  &  \\
\bottomrule
\end{tabular}
\vspace{-1em}
\end{table}

The comparison results are reported in Table~\ref{T:kohler}. Similarly to GoPro, SRN and DeblurGAN-v2 (Inception-ResNet-v2) remain to be the best two PSNR/SSIM performers, but this time SRN is marginally superior in both. However, please be reminded that, similarly to the GoPro case, this ``almost tie'' result was achieved while DeblurGAN-v2 (Inception-ResNet-v2) costs only 1/5 of SRN's inference complexity. Moreover, both DeblurGAN-v2 (MobileNet) and DeblurGAN-v2 (MobileNet-DSC) outperform DeblurGAN on the Kohler dataset in both SSIM and PSNR: that is impressive given the former two's much lighter weights.

Figure \ref{fig:kohler_test} displays visual examples on the Kohler dataset. DeblurGAN-v2 effectively restores the edges and textures, without noticeable artifacts. SRN for this specific example shows some color artifacts when zoomed in. 


\subsection{Quantitative Evaluation on DVD dataset}
\vspace{-0.5em}

\begin{table*}[htbp] \scriptsize
\vspace{-1em}
	\caption{Average subjective scores of deblurring results on the Lai dataset~\cite{lai2016comparative}.}
    \vspace{-1.5em}
	\begin{center}\small{
			\begin{tabular}{|c|c|c|c|c|c|c}
				\hline
			   Blurry & Krishnan~\etal \cite{normalized_sparsity} & Whyte~\etal~\cite{whyte} & Xu~\etal ~\cite{XuUnnaturalDeblurring}  & Sun~\etal~\cite{SunLearningRemoval}&  Pan~\etal~\cite{darkchannelprior} \\
			   \hline
			   1 & 1.08 & 0.57 & 0.77 & 0.64 & 0.91 \\
			   \hline
			   \hline
			   DeepDeblur \cite{Nah2016DeepDeblurring}  & SRN \cite{tao2018scale}  &  DeblurGAN \cite{kupyn2018deblurgan}  & DeblurGAN-v2  & DeblurGAN-v2 &  DeblurGAN-v2  \\
			   	 & &    & (Inception-ResNet-v2)  &  (MobileNet) &  (MobileNet-DSC) \\
			   \hline
			   1.08 & 1.68 & 1.29 & \textbf{1.74} & 1.44 & 1.32 \\
				\hline
		\end{tabular}}
		\label{tab-subjective}
	\end{center}
	\vspace{-1em}
\end{table*}

We next test DeblurGAN-v2 on the DVD testing set used in \cite{su2017deep}, but with a \textit{single-frame} setting (treating all frames as individual images) without using multiple frames together. We compare with two strong video deblurring methods: WFA \cite{delbracio2015burst}, and DVD \cite{su2017deep}, 
For the latter, we adopt the authors' self-reported results when using a single frame as the model input (denoted as ``single''), for a fair comparison. As shown in Table \ref{DVD}, DeblurGAN-v2 (MobileNet) outperforms WFA and DVD (single), while being at least 17 times faster (DVD was tested on a reduced resolution of 960 $\times$ 540, while DeblurGAN-v2 is on 1280 x 720). 

While not specifically optimized for video deblurring, DeblurGAN-v2 shows good potential, and we will extend it to video deblurring as future work.



\vspace{-0.2em}
\subsection{Subjective Evaluation on Lai dataset}
\vspace{-0.5em}
The Lai dataset \cite{lai2016comparative} has real-world blurry images of different qualities and resolutions collected in various types of scenes. Those real images have no clean/sharp counterparts, making a full-reference quantitative evaluation impossible. Following \cite{lai2016comparative}, we conduct a subjective survey to compare the deblurring performance on those real images. 

We fit a Bradley-Terry model \cite{bradley1952rank} to estimate the subjective score for each method so that they can be ranked, with the identical routine following the previous benchmark work \cite{li2019benchmarking,li2019single}. Each blurry image is processed with each of the following algorithms: Krishnan~\etal \cite{normalized_sparsity}, Whyte~\etal~\cite{whyte}, Xu~\etal ~\cite{XuUnnaturalDeblurring}, Sun~\etal~\cite{SunLearningRemoval}, Pan~\etal~\cite{darkchannelprior}, DeepDeblur \cite{Nah2016DeepDeblurring}, SRN \cite{tao2018scale}, DeblurGAN \cite{kupyn2018deblurgan}; and the three DeblurGAN-v2 variants (Inception-ResNet-v2, MobileNet, MobileNet-DSC). The eleven deblurring results, together with the original blurry image, are sent for pairwise comparison to construct the winning matrix. We collect the pair comparison results from 22 human raters. We observed good consensus and small inter-person variances among raters, which makes scores reliable. 

The subjective scores are reported in Table \ref{tab-subjective}. We did not normalize the scores due to the absence of ground-truth: as a result, it is the score rank rather than the absolute score value that matters here. It can be observed that deep learning-based deblurring algorithms, in general, have more favorable visual results than traditional methods (some even making visual quality worse than the blurry input). DeblurGAN \cite{kupyn2018deblurgan} outperforms DeepDeblur \cite{Nah2016DeepDeblurring}, but lags behind SRN \cite{tao2018scale}. With the Inception-ResNet-v2 backbone, DeblurGAN-v2 demonstrates clearly superior perceptual quality over SRN, making it the top performer in terms of subjective quality. DeblurGAN-v2 with MobileNet and MobileNet-DSC backbones have minor performance degradations compared to the Inception-ResNet-v2 version. However, both are still preferred by subjective raters, compared to DeepDeblur and DeblurGAN, while being 2-3 orders-of-magnitude faster. 

Figure \ref{fig:lai} displays visual comparison examples on deblurring the ``face2'' image. DeblurGAN-v2 (Inception-ResNet-v2) (\ref{fig:lai_our}) and SRN (\ref{fig:lai_k}) are the top-2 most favored results, both balancing well between edge-sharpness and overall smoothness. By zooming in, SRN is found to still generate some ghost artifacts on this example, e.g., the white ``intrusion'' from the collar to the bottom right face region. In comparison, DeblurGAN-v2 (Inception-ResNet-v2) shows artifact-free deblurring. Besides, DeblurGAN-v2 (MobileNet) and DeblurGAN-v2 (MobileNet-DSC) results are also smooth and visually better than DeblurGAN, though less sharper than DeblurGAN-v2 (Inception-ResNet-v2).


\subsection{Ablation Study and Analysis}
\vspace{-0.5em}
We perform an ablation study on the effect of specific components of the DeblurGAN-v2 pipeline. Starting from the original DeblurGAN (ResNet G, local-scale patch D, WGAN-GP + perceptual loss), we gradually inject our modifications on the generator (adding FPN), discriminator (adding global-scale), and the loss (replacing WGAN-GP loss with RaGAN-LS, and adding an MSE term). The results are summarized in Table \ref{T:restore}. We can see that all our proposed components steadily improve both PSNR and SSIM. In particular, the FPN module contributes most significantly. Also, adding either MSE or perceptual loss benefits both training stability and final results.


\begin{table}[htb]
\small
\vspace{-0.5em}
\caption{Ablation Study on the GoPro dataset, based on DeblurGAN-v2 (Inception-ResNet-v2).}
\label{T:restore}
\centering
\setlength\tabcolsep{3pt}
\begin{tabular}{c|c|c}
\toprule
 & PSNR & SSIM \\
\hline
DeblurGAN (starting point) & 28.70 & 0.927  \\
\hline
+ FPN & 29.26  & 0.931 \\
\hline
+ FPN + Global D & 29.29 & 0.932 \\
\hline
+ FPN +  Global D + RaGAN-LS & 29.37 & 0.933 \\
\hline
\textbf{DeblurGAN-v2} (FPN + Global D +  & & \\
RaGAN-LS + MSE Loss) & \textbf{29.55} & \textbf{0.934} \\
\hline
Removing perceptual loss &  &  \\
(replace 0.5 with 0 in $L_G$) & 28.81  & 0.924 \\
\bottomrule
\end{tabular}
\vspace{-0.5em}
\end{table}
As an extra baseline for the efficiency of FPN, we tried to create a ``compact'' version of SRN, with roughly the same FLOPs (456 GFLOPs) to match DeblurGAN-v2 Inception-ResNet-v2 (411 GFLOPs). We reduced the numbers of ResBlocks by 2/3 in each EBlock/DBlock while keeping their 3-scale recurrent structure. We then compare with DeblurGAN-v2 (Inception-ResNet-v2) on GoPro, where that ``compact'' SRN only achieved PSNR = 28.92 dB and SSIM = 0.9324. We also tried channel pruning \cite{he2017channel} to reduce SRN FLOPs and the result was no better. 

\begin{table}[htb]
\small
\caption{PSNR/SSIM comparison on Restore Dataset.}
\vspace{-1em}
\label{T:restore}
\centering
\begin{tabular}{c|cc}
\toprule
& PSNR & SSIM \\
\hline
 Degraded & 22.056 &  0.873 \\
DeblurGAN & 26.435 &  0.892 \\
DeblurGAN-v2 (Inception-ResNet-v2) & \textbf{26.916} &  \textbf{0.894} \\
DeblurGAN-v2 (MobileNet-DSC) & 25.412 & 0.891 \\
\bottomrule
\end{tabular}
\vspace{-1.2em}
\label{DVD}
\end{table}

\subsection{Extension to General Restoration}
\vspace{-0.4em}
Real-world atural images commonly go through multiple kinds of degradations (noise, blur, compression, etc.) at once, and a few recent works were devoted to such join enhancement tasks \cite{lsd2,gated}
We study the effect of DeblurGAN-v2 on the task of general image restoration. While \textbf{NOT} being the main focus of this paper, we intend to show the general architecture superiority of DeblurGAN-v2, especially for modifications made w.r.t. DeblurGAN. 

We synthesize a new challenging \textit{Restore Dataset}. We take 600 images from GoPRO, and 600 images from DVD, both with motion blurs already (same as above). We then use the \textit{albumentations} library \cite{2018arXiv180906839B} to further add Gaussian and speckle Noise, JPEG compression, and up-scaling artifacts to those images. Eventually, we split 8000 images for training and 1200 for testing. We train and compare DeblurGAN-v2 (Inception-ResNet-v2), DeblurGAN-v2 (MobileNet-DSC), and DeblurGAN. As shown in Table~\ref{T:restore} and Fig. \ref{fig:yolo}, 
DeblurGAN-v2 (Inception-ResNet-v2) achieves the best PSNR, SSIM, and visual quality. 



\vspace{-0.5em}
\section{Conclusion}
\vspace{-0.5em}
This paper introduces DeblurGAN-v2, a powerful and efficient image deblurring framework, with promising quantitative and qualitative results. DeblurGAN-v2 enables to switch between different backbones, for flexible trade-offs between performance and efficiency. We plan to extend DeblurGAN-v2 for real-time video enhancement, and for better handling mixed degradations.

\textbf{Acknowledgements:}
O. Kupyn and T. Martyniuk were supported by SoftServe, Let's Enhance, and UCU. J. Wu and Z. Wang were supported by NSF Award RI-1755701. The authors thank Arseny Kravchenko, Andrey Luzan and Yifan Jiang for constructive discussions, and Igor Krashenyi and Oles Dobosevych for computational resources.


{\small
\bibliographystyle{ieee_fullname}
\bibliography{deblurgan_v2}
}
\end{document}